# Accurate but Natural? Diagnosing Grammatical and Idiomatic Gaps in Japanese EFL Writing

**Steve Woollaston[1], Brendan Flanagan[2], Hiroaki Ogata[1]**
[1]Kyoto University, [2]Ritsumeikan University
[1]Academic Center for Computing and Media Studies, Kyoto University,
[2]College of Information Science and Engineering, Ritsumeikan University
s.m.woollaston@gmail.com

**Abstract**
Second language writing research distinguishes grammatical accuracy from native-like idiomaticity, yet automated writing evaluation often conflates these dimensions. This study introduces a layered LLM-correction pipeline that isolates structural errors from unnaturalness by generating literal error corrections and idiomatic revisions for 3,830 English writing samples from 120 Japanese junior high school students. Applying the regex-based CEFR-J grammar extractor, we quantify two diagnostic measures: accuracy gaps (structures attempted but incorrectly produced) and idiomatic gaps (grammatically correct structures underused or overused relative to native norms). Results reveal distinct patterns: definite articles, third-person singular *-s*, and modals (*would*, *could*) exhibit significant accuracy difficulties, while *-ing* forms and hypothetical modals (*would*) show the largest idiomatic underuse, with simple present verbs, subject-verb-object patterns, and modal *can* conversely exhibiting the most pronounced overuse. A two-dimensional instructional typology maps error rates against idiomatic gaps, distinguishing accurate but overused grammar items from error-prone or avoided complex forms requiring targeted production practice. This framework advances pedagogical feedback by enabling teachers to diagnose whether learner difficulties arise from inaccurate execution, structural avoidance, or L1-mapped overreliance, supporting evidence-based interventions tailored to the specific needs of each learner.

## 1. Introduction

Second language (L2) writing research distinguishes grammatical accuracy from native-like (L1) idiomaticity. A text may be error-free yet sound unnatural due to L1 transfer (Granger, 2015; Jarvis & Pavlenko, 2010). This distinction is especially pronounced for Japanese learners of English, given the linguistic distance across article systems, verb morphology, clause structure, and pragmatic marking (Ishii & Tono, 2018; Muroya, 2018; Takahashi, 2016). Yet Automated Writing Evaluation (AWE) and learner corpus approaches have historically prioritised grammatical accuracy, which is easier to detect and count, over stylistic naturalness, which is more difficult to define. Yorio (1989) observed that grammaticality is readily identified by the absence of formal error, while idiomaticity remains elusive and reflects conventionalised rather than rule-governed language. He further noted that fluency can exist without full grammatical accuracy, but idiomaticity requires both formal correctness and appropriate language-specific usage, making it a qualitatively higher bar. Consequently, these frameworks gauge output against a narrow benchmark, treating divergences as undifferentiated errors rather than separating structural inaccuracy from L1-induced unnaturalness (Götz & Granger, 2024; Granger & Rayson, 1998). This conflation obscures distinct pedagogical needs: grammar instruction for accuracy versus naturalistic input for idiomaticity.

LLMs offer a pathway, producing two versions of the same text: one minimally corrected to preserve structure and one naturalised to reflect native-like phrasing (Coyne et al., 2023; Davidson, 2024). This enables isolation of the accuracy gap, representing attempted but erroneous structures, from the idiomatic gap, representing correct but L1-constrained phrasing, as shown in Figure 1. This study introduces and validates this framework on Japanese junior high school EFL writing. By comparing raw uncorrected, literal-corrected, and idiomatically revised versions, we quantify both gaps and generate diagnostic profiles, moving pedagogical intervention from intuition to evidence.

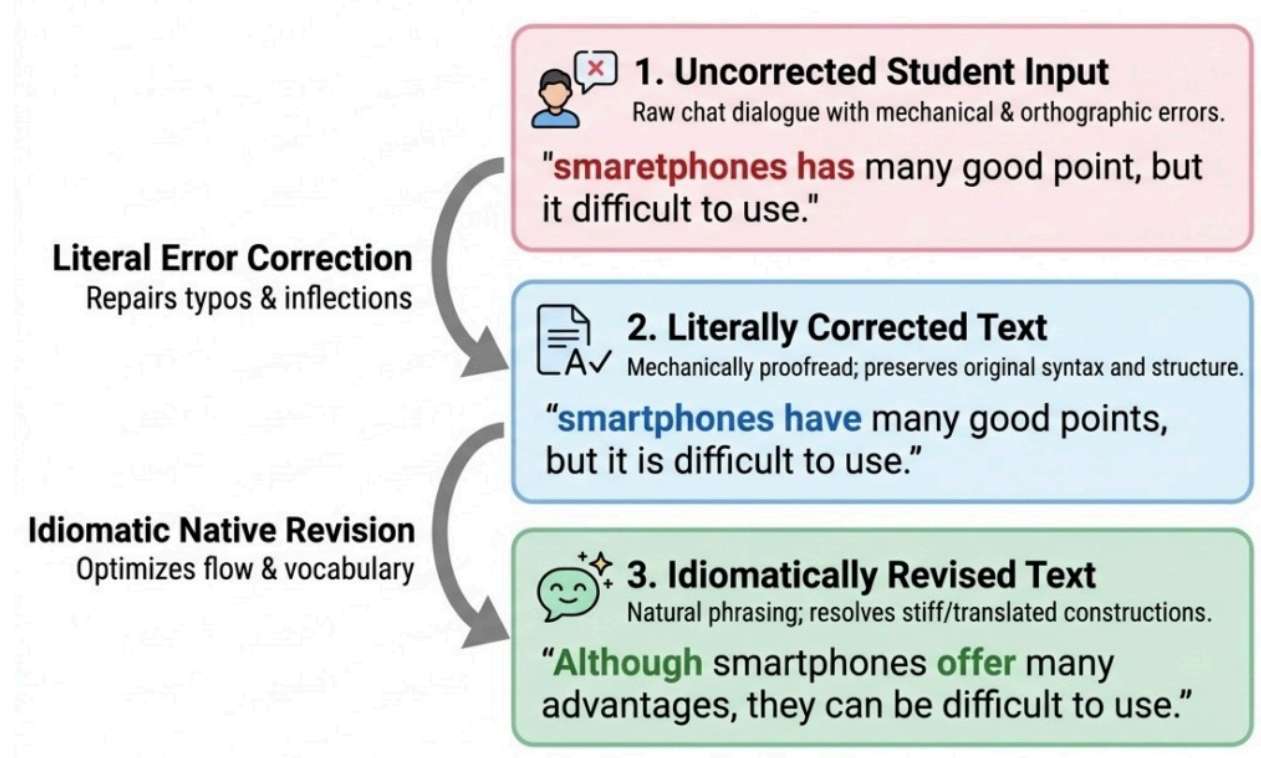

Figure 1: Accuracy vs. Idiomatic Gap.

This paper addresses the following research questions:

1. Which grammatical structures exhibit the largest **accuracy gaps** (i.e., largest frequency changes from uncorrected-raw to literal-corrected versions), indicating that learners attempt them frequently but fail to execute them correctly?
2. Which grammatical structures show the widest **idiomatic gaps** (i.e., largest frequency changes from literal-corrected to idiomatic-naturalised versions), revealing L1-induced constraints on natural phrasing even in error-free writing?
3. How can grammatical structures be systematically categorised into distinct **instructional priorities** based on their dual profiles of production accuracy (accuracy gap) and frequency of use compared to native norms (idiomatic gap)?

## 2. Related Work

### 2.1 CEFR-J Grammar Profile

A key framework for diagnosing Japanese EFL writing development is the CEFR-J (Tono, 2013), an adaptation of the Common European Framework of Reference optimised for English language education in Japan. Beyond providing fine-grained 'Can-Do' descriptors, the framework includes the CEFR-J Grammar Profile (Ishii & Tono, 2016), which maps 501 specific grammar structures across language proficiency tiers (A1.1 to B2.2). Because Japanese junior high school learners operate primarily at the lower CEFR tiers, the CEFR-J's expanded subdivisions at the A1 and A2 levels provide the diagnostic sensitivity necessary to track low-frequency or developing grammatical features (see cefr-j.org for more information).

## 2.2 Accuracy Profiles and L1 Transfer

Japanese EFL writing shows clear patterns of error. Quantitative corpus profiling has become a useful method for diagnosing these patterns by comparing learner output against native-corrected parallel texts (Granger & Rayson, 1998). Ishii and Tono (2018) built a comprehensive grammar profile of CEFR-J grammar patterns by comparing essays from the Japanese EFL Learner (JEFLL) corpus with native-corrected versions. Their work revealed systematic underuse (omission of obligatory structures) at lower proficiency, and overuse of simple L1-mapped defaults. For instance, the lack of an article system in Japanese leads to a 46% article omission rate at the A1 level, and standard instruction translating the as *sono* (that) and "*a"* as *hitotsuno* (one) further restricts semantic interpretation (Hinenoya & Lyster, 2015; Ishii & Tono, 2018). By contrast, the possessive *'s* is acquired early and accurately because it closely matches the Japanese particle *-no* (Luk & Shirai, 2009; Murakami & Alexopoulou, 2016). Another persistent issue is the overuse of *be*, as in "It is ramen" instead of "I eat ramen." This comes from mapping the Japanese topic-comment pattern directly onto English (Ishii & Tono, 2018). Verb endings also reveal L1 effects. Regular past tense *-ed* is relatively easy for Japanese learners because it resembles the Japanese past marker *-ta*, while the third-person singular *-s* (e.g., *he plays*) is much harder—it packs person, number, and tense into one small ending, something Japanese does not require (Muroya, 2018). This reverses the typical accuracy order found in learners from other L1 backgrounds.

Even when grammar is flawless, Japanese learners' English writing often lacks native-like flow. Phrasal verbs (*give up*, *run out of*) are used far less than expected because Japanese expresses motion and result inside the main verb, without adding directional particles like *up* or *out* (Strong et al., 2026). Aspect is another area of transfer. The Japanese *-te iru* form can signal an ongoing action, a habit, or a resulting state. Learners therefore overextend English *-ing* to verbs like *know* and *love* ("he is knowing") and, at other times, drop it where it belongs, relying on time words like "now" instead (e.g., "He studies Japanese now", "She eats breakfast now") (Bryant, 1984).

## 2.3 LLMs for Correction and Idiomatic Revision

Manual correction of learner writing is slow, inconsistent, and often mixes grammatical fixes with style changes (Götz & Granger, 2024; Takahashi, 2016). LLMs now offer a fast, scalable way to correct grammar and to rewrite for fluency, acting as a "copilot" that helps non-native writers sound more natural (Coyne et al., 2023; Davidson, 2024; Torrent et al., 2023). The risk is that LLMs can over-correct, rewriting more than necessary and losing the learner's original structure (Fang et al., 2023; Zeng et al., 2024). However, when prompts are carefully designed, these tools can bridge the gap between getting it right and sounding natural (Davidson, 2024).

# 3. Methodology

## 3.1 Participants and Setting

The raw writing samples (3,912 total) were collected from 120 Japanese junior high school students (aged 14-16) across three EFL classes. Over a four-month period (November 2024 – March 2025), learners used a writing support chatbot daily during English lessons (3-4 sessions per week). Each writing session followed a five-minute "Think-Pair-Share" activity followed by 10-20 minutes of independent writing and chatbot interaction with feedback. After cleaning blank and nonsensical writing, the corpus comprises 3,830 writing submissions, all responses to a variety of diary prompts (e.g., *What is your least favourite food and why?*; *Describe a place you've visited recently.*). The full question list is available on request.

## 3.2 System

The system, named *Penny* (Figure 2), was developed by the authors in collaboration with a local teacher. It provided a different writing prompt each day and accepted only English text. Interaction with the chatbot for feedback was provided when the learner clicked the ***Check my writing*** button, which was disabled until the learner had written at least 100 characters, ensuring that all submissions were unassisted first drafts. *Penny* was powered by OpenAI's GPT-4o (`gpt-4o-2024-08-06`) and was accessible at any time, with a teacher dashboard for monitoring progress and engagement. All interactions were logged, including chatlogs, revision histories, and button clicks. Chatbot interactions and subsequent learner revisions were excluded from the present analysis (see Woollaston et al., 2026).

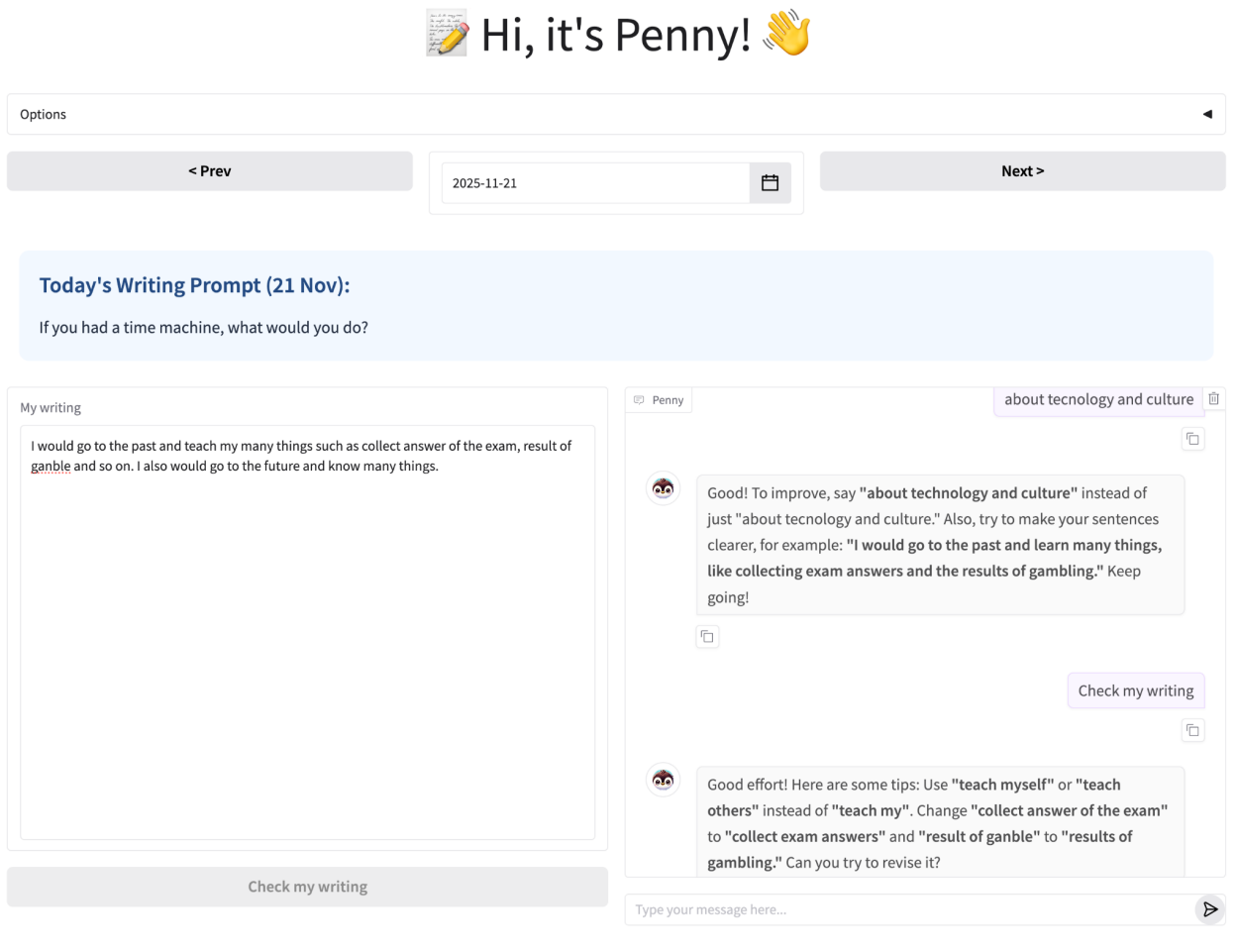

Figure 2: Penny system interface.

## 3.3 Learner Text Processing

To isolate the **accuracy gap** (structures attempted but executed with errors) and the **idiomatic gap** (grammatically correct but L1-constrained phrasing), we generated two versions of each raw writing submission using an LLM:

1. **Literal Error Correction**: Generated using a prompt to fix only spelling, punctuation, and mechanical grammatical errors (e.g., subject-verb agreement, tense, articles) without altering sentence structure or word order (`temp = 0.1`).

2. **Idiomatic Natural Revision**: Generated using a prompt to rewrite the text into more natural, native-like English while preserving the core meaning, with freedom to adjust vocabulary, syntax, and idiomatic expressions (`temp = 0.3`).

A local LLM (`gemma4:31b`) was used for text processing to ensure reproducibility and data privacy, mitigating reliance on proprietary APIs whose models and endpoints are subject to undocumented deprecation or change. The capacity of modern LLMs to reliably repair surface grammar and rephrase texts naturally is well-documented in recent literature (Lin et al., 2026; Wu et al., 2023). Latest-generation foundational models consistently outperform standard rule-based commercial grammar checkers in error detection, grammatical correction, and naturalness preservation. Furthermore, when evaluated beyond rigid reference-based string matching, modern LLMs generate fluent, native-like revisions that human annotators rate as valid or superior in the vast majority of cases (Labib et al., 2026).

### 3.4 Semantic Fidelity and Validation

To ensure the layered text-processing pipeline remained faithful to student intent without introducing hallucinations or semantic drift, we evaluated semantic similarity across all three text layers using text embeddings (`embeddinggemma-300m`). Cosine similarity was calculated across the entire corpus (N = 3,830) for all layers: Uncorrected vs. Literal, Uncorrected vs. Idiomatic, and Literal vs. Idiomatic.

| Layers | Cosine Similarity (Mean ± SD) | Median | IQR | Min / Max |
|---|---|---|---|---|
| **Uncorrected vs. Literal** | 0.968 ± 0.041 | 0.980 | 0.959 – 0.991 | 0.105 / 1.000 |
| **Uncorrected vs. Idiomatic** | 0.843 ± 0.098 | 0.865 | 0.810 – 0.906 | 0.084 / 0.992 |
| **Literal vs. Idiomatic** | 0.865 ± 0.096 | 0.888 | 0.835 – 0.924 | 0.052 / 1.000 |

Table 1: Cosine similarity distributions across layers

As shown in Table 1, the Literal Correction layer strictly preserved the original content, with 94.44% of samples achieving $\text{Similarity}_{\text{Literal}} \geq 0.90$ and minimal length variation ($\Delta$ = +1.00 words per session). This confirms that mechanical corrections did not alter underlying syntax or propositions. The Idiomatic Revision layer balanced natural restructuring with meaning preservation, with 77.89% achieving $\text{Similarity}_{\text{Idiomatic}} \geq 0.80$ when compared to the uncorrected submissions. The expected moderate decrease in cosine similarity and the 1.26x mean expansion ratio reflect natural grammar revision and lexical diversification (mean new vocabulary introduction rate of 63.42%). Table 2 provides two illustrative examples of the text layer processing and their cosine similarities.

| Student Text and Derivatives | Cosine Similarity |
|---|---|
| **Prompt:** What is the silliest thing you've ever done? | |
| *Uncorrected:* “My silliest thing is broken my glasses. I pulled my glasses with a chair. I god scolded by my mom. I reflected. but I broke it again.” | |
| *Literal:* “The silliest thing is I broke my glasses. I pulled my glasses with a chair. I got scolded by my mom. I reflected, but I broke them again.” | 0.959 |
| *Idiomatic:* “The silliest thing I've ever done was break my glasses. I managed to snag them on a chair and pull them right off my face. My mom gave me a real earful about it, and I promised her I'd be more careful. I really thought I had learned my lesson, but then I went and broke them all over again.” | 0.904 |
| **Prompt:** If you could travel back in time, what time period would you visit? | |
| *Uncorrected:* “I want to go Edo period with smartphone. Because I want to take over the world to use CP and AI. Mayby I can it, if I am storong.” | |
| *Literal:* “I want to go to the Edo period with a smartphone. Because I want to take over the world using PC and AI. Maybe I can do it, if I am strong.” | 0.938 |
| *Idiomatic:* “I would travel back to the Edo period, and I'd definitely bring a smartphone with me. My goal would be to take over the world by introducing computers and AI. I feel like if I played my cards right and stayed strong, I could actually pull it off.” | 0.850 |

Table 2: Examples of processed text layers

To assess the extreme lower bound of semantic preservation, we conducted a targeted audit of all sessions exhibiting cosine similarity below 0.5 across any layer comparison. Out of 3,830 total sessions, only 55 (1.44%) fell below this threshold, with the vast majority of divergences occurring between the literal and idiomatic layer (51 sessions), while uncorrected versus literal preservation remained virtually unaffected (only 2 sessions < 0.5). Qualitative inspection revealed that these low-scoring outliers did not stem from pipeline errors on valid student writing; rather, 100% of the flagged records represented invalid user inputs and edge cases. These fell into distinct categories: repetition (presumably to reach the minimum character threshold and chat with Penny) and keyboard spam (e.g., repeated character strings, memes, or profanity), input in Romanised Japanese (rōmaji) rather than English, insults, and system refusals. In most of these scenarios, the literal layer correctly preserved the raw text mechanically, whereas the idiomatic model appropriately triggered conversational safety refusals, meta-feedback requesting a valid draft, or contextual English translations of the Romanised text, naturally yielding low vector alignment. The absence of low-similarity anomalies on legitimate student submissions confirms that semantic fidelity remained intact throughout the corpus.

### 3.5 Grammar Analysis

All versions were processed by the CEFR-J regex grammar extractor (Ishii & Tono, 2016), a rule-based system chosen to avoid LLM hallucinations. After normalising frequencies per 10,000 words, we retained only features with a minimum frequency of at least five in the target layer to reduce noise from rare occurrences. We defined the **accuracy gap** as $Freq_{literal} - Freq_{raw}$ and the **idiomatic gap** as $Freq_{idiomatic} - Freq_{literal}$. To assess shifts between layers, we used Dunning's Log-Likelihood ($G^2$; Dunning, 1993), standard for non-normal, low-frequency data (Gabrielatos, 2018). Due to large N, we paired *p*-values with Hardie's (2014) log ratio:

$$Log\ Ratio = log_2\left(\frac{Norm.Freq_{Layer\ B} + 0.1}{Norm.Freq_{Layer\ A} + 0.1}\right)$$

We apply a smoothing constant ($\varepsilon = 0.1$) to avoid undefined logs while preserving signal. Log ratio thresholds of $\geq 1$ and $\geq 2$ indicate doubling/halving and quadrupling/quartering, respectively. Reporting raw density shifts ($\Delta$) alongside $G^2$ maximises interpretability for educational practitioners, while log ratio captures relative proportional changes, effectively distinguishing broad volume shifts from structural intensity (the baseline trap).

## 4. Results

To enable valid comparisons across the three text versions, we computed the total word count, number of sentences, and words per sentences for each layer (Table 3). The Literal Correction layer was slightly longer than the Uncorrected layer, while the Idiomatic Revision layer expanded substantially as the LLM introduced more explicit connectors, elaborated ideas, and added cohesive devices to achieve more native-like fluency.

| Layer | Words | Sentences | Words / Sentence |
|---|---|---|---|
| Uncorrected | 150,654 | 18,562 | 8.12 |
| Literal Correction | 154,478 | 18,871 | 8.19 |
| Idiomatic Revision | 182,671 | 12,998 | 14.05 |

Table 3: Corpus counts by layer

All subsequent frequency reports are normalised to occurrences per 10,000 words to account for differences in text volume.

### 4.1 RQ1: Accuracy Gap

To identify which grammatical structures learners attempt but execute with sufficient errors to render them unrecognisable to the grammar extractor, we compared normalised frequencies between the Uncorrected and Literal Correction layers. Positive values indicate structures that increased in detectability after mechanical errors were repaired. Table 4 presents the top 15 frequency shifts from Uncorrected to Literal Correction.

| Shortcode | Grammar Item and Example | CEFR Range | Raw Norm | Literal Norm | Shift (Δ) | $G^2$ | Log Ratio |
|---|---|---|---|---|---|---|---|
| DT.the | DEFINITE ARTICLES (e.g., *the book*) | A1.1 | 97.31 | 114.58 | +17.27 | 21.72*** | 0.24 |
| TA.PRESENT.do.AFF | TENSE/ASPECT: PRESENT (lexical verbs) (e.g., *I play*) | A1.1 | 208.96 | 197.37 | –11.59 | 5.14* | –0.08 |
| TA.PRESENT.be.AFF | TENSE/ASPECT: PRESENT (BE) (e.g., *I am*) | A1.1 | 191.17 | 184.36 | –6.81 | $1.92^{ns}$ | –0.05 |
| TA.PAST.do.AFF | TENSE/ASPECT: PAST (lexical verbs) (e.g., *I played*) | A1.2–A1.3 | 52.97 | 58.97 | +6.00 | 4.94* | 0.15 |
| TA.PRESENT.does.AFF | TENSE/ASPECT: PRESENT (3rd person singular) (e.g., *he plays*) | A1.1–A1.2 | 18.32 | 23.24 | +4.92 | 8.92** | 0.34 |
| VG | V-ING (not preceded by 'not') (e.g., *I like swimming, He is eating*) | A1.2–A2.2 | 87.75 | 92.51 | +4.76 | $1.93^{ns}$ | 0.08 |
| CL_after.etc | SUBORDINATE CLAUSE (e.g., *because I was tired*) | A1.3–B1.2 | 55.96 | 59.36 | +3.40 | $1.54^{ns}$ | 0.09 |
| MD.would.AFF | MODAL/AUX: would (e.g., *I would go*) | A1.3–B1.2 | 5.71 | 8.61 | +2.90 | 9.03** | 0.58 |
| PP.they_are | they are (e.g., *They are happy*) | A1.1–A1.2 | 8.56 | 11.26 | +2.70 | 5.63* | 0.39 |
| MD.could.AFF | MODAL/AUX: could (e.g., *I could help*) | A2.1–B1.2 | 4.05 | 6.54 | +2.49 | 9.01** | 0.68 |
| VP.SVO.AFF | SENTENCE PATTERN: SUBJECT+V+OBJECT (e.g., *I like dogs*) | A1.1–A2.2 | 36.44 | 38.91 | +2.47 | $1.23^{ns}$ | 0.09 |
| TA.PAST.be.AFF | TENSE/ASPECT: PAST (BE) (e.g., *I was*) | A1.2–A1.3 | 38.43 | 40.65 | +2.22 | $0.95^{ns}$ | 0.08 |
| MD.could.NEG | MODAL/AUX: could (negative) (e.g., *could not go*) | A2.1–B1.2 | 1.53 | 3.69 | +2.16 | 14.08*** | 1.22 |
| TA.PAST.do.NEG | TENSE/ASPECT: PAST (lexical verbs, negative) (e.g., *did not go*) | A1.2–A1.3 | 2.85 | 4.98 | +2.13 | 8.94** | 0.78 |

| Shortcode | Grammar Item and Example | CEFR Range | Raw Norm | Literal Norm | Shift (Δ) | $G^2$ | Log Ratio |
|---|---|---|---|---|---|---|---|
| CL.when | ADVERBIAL CLAUSE: when (e.g., *when I arrived*) | A1.2 | 22.37 | 20.33 | –2.04 | $1.49^{ns}$ | –0.14 |

Table 4: Top 15 Accuracy Gap Shifts. *Significance levels: *** p < .001, ** p < .01, * p < .05, ns = not significant.*

Among the significant positive shifts, definite articles (**DT.the**) stand out with the largest absolute gain. Three other structures show substantial increases with moderate significance: past tense lexical verbs **(TA.PAST.do.AFF**), third-person singular present (**TA.PRESENT.does.AFF**), and the pronoun + BE construction they are (**PP.they_are**). The modal category presents a notable concentration of significant gaps. Both affirmative forms of would (**MD.would.AFF**) and could (**MD.could.AFF**) show robust positive shifts, while the negative form could (**MD.could.NEG**) yields the largest Log ratio among all accuracy shifts, reflecting a marked proportional increase from a low baseline. The negative past tense construction **TA.PAST.do.NEG** also shows a significant positive shift.

Several positive shifts did not reach statistical significance. These include V-ING (**VG**), subordinate clauses (**CL_after.etc**), SVO sentence patterns (**VP.SVO.AFF),** and past tense BE (**TA.PAST.be.AFF**), all of which show modest increases but lack sufficient confidence to rule out chance variation. The only significant negative shift in the top 15 is present tense lexical verbs (**TA.PRESENT.do.AFF**). Two other features, present tense BE (**TA.PRESENT.be.AFF**) and adverbial clauses with *when* (**CL.when**), also decreased, though neither reached statistical significance.

## 4.2 RQ2: Idiomatic Gap

To identify grammatical structures that differ in frequency between error-free learner writing and native-like revisions, we compared normalised frequencies between the Literal Correction and Idiomatic Revision layers. In the **idiomatic gap**, positive values indicate structures that increased in the native-like revision and negative values indicate structures that decreased.

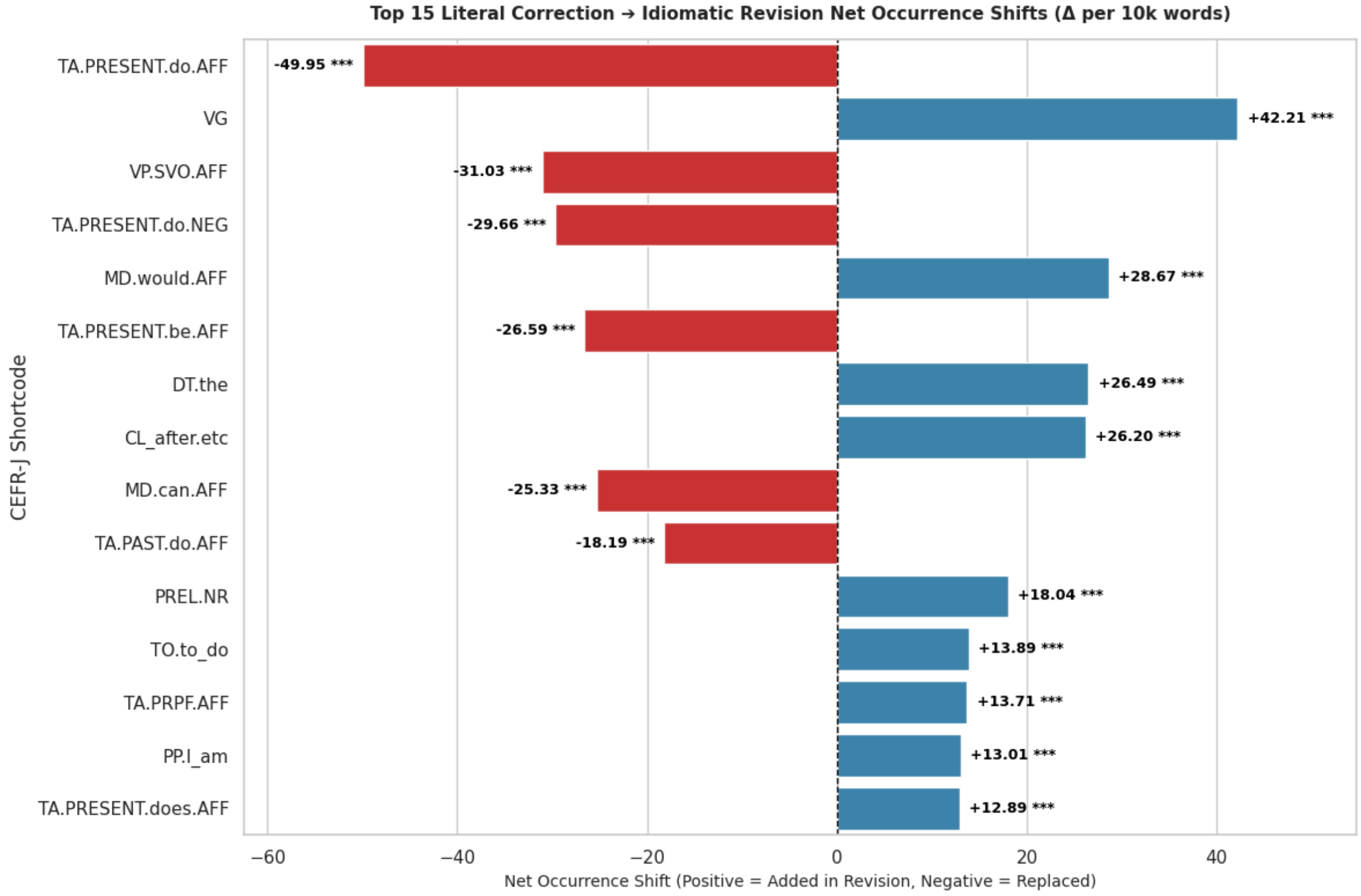

Figure 3: Largest frequency shifts from Literal Correction to Idiomatic Revision. Significance level: *** $p < .001$.

As shown in Figure 3, the largest positive shift was observed for V-ING (**VG**), which increased by +42.21 per 10k words ($G^2$ = 132.90, log ratio = 0.54). The modal would (**MD.would.AFF**) showed the second-largest increase of +28.67 ($G^2$ = 318.11, log ratio = 2.10). Definite articles **(DT.the**) increased by +26.49 ($G^2$ = 46.51, log ratio = 0.30), and subordinate clauses (**CL_after.etc)** increased by +26.20 ($G^2$ = 79.92, log ratio = 0.53). Nonrestrictive relative pronouns (**PREL.NR**) exhibited the largest log ratio among all idiomatic shifts at 3.86, corresponding to an increase of +18.04 ($G^2$ = 311.43, $p < .001$).

To-infinitives (**TO.to_do**) increased by +13.89 ($G^2$ = 11.57, log ratio = 0.14), while present perfect (**TA.PRPF.AFF**) showed a large increase of +13.71 ($G^2$ = 114.35, log ratio = 1.51). I am (**PP.I_am**) increased by +13.01 ($G^2$ = 28.37, log ratio = 0.38), and present third-person singular (**TA.PRESENT.does.AFF**) increased by +12.89 ($G^2$ = 47.08, log ratio = 0.63).

Present tense lexical verbs (**TA.PRESENT.do.AFF**) showed the largest negative shift of –49.95 ($G^2$ = 124.18, log ratio = –0.42). SVO sentence patterns (**VP.SVO.AFF**) decreased by –31.03 ($G^2$ = 383.91, log ratio = –2.29), representing the largest log ratio magnitude among all idiomatic shifts. Present tense negative forms (**TA.PRESENT.do.NEG**) decreased by –29.66 ($G^2$ = 266.53, log ratio = –1.59), and present tense BE (**TA.PRESENT.be.AFF**) decreased by –26.59 ($G^2$ = 35.31, log ratio = –0.22). The modal can (**MD.can.AFF**) decreased by –25.33 ($G^2$ = 152.12, log ratio = –1.04), and past tense lexical verbs (**TA.PAST.do.AFF**) decreased by –18.19 ($G^2$ = 56.39, log ratio = –0.53).

## 4.3 RQ3: Instructional Typology of Grammatical Targets

To answer RQ3, the top 40 grammatical targets, selected by overall volume across all three corpus layers ($Uncorrected + Corrected + Idiomatic$) were mapped across two diagnostic axes: **Error Rate (%)** and **Idiomatic Gap**. The size of each node in the resulting scatter plot represents the target's overall frequency across the dataset. To determine statistical significance across both dimensions, Log-Likelihood ($G^2$) tests (α = 0.05) were calculated for shifts between corpus layers. Rather than using absolute frequency changes for structural accuracy, horizontal position reflects the relative error rate calculated as:

$$Error\ Rate\ (\%) \; = \frac{Corrected_{raw} - Uncorrected_{raw}}{Corrected_{raw}} \times 100.$$

This controls for base frequency so that high-volume default structures do not artificially mask the difficulty of rarer forms. The vertical position reflects the **idiomatic gap**, measured as the normalised frequency shift.

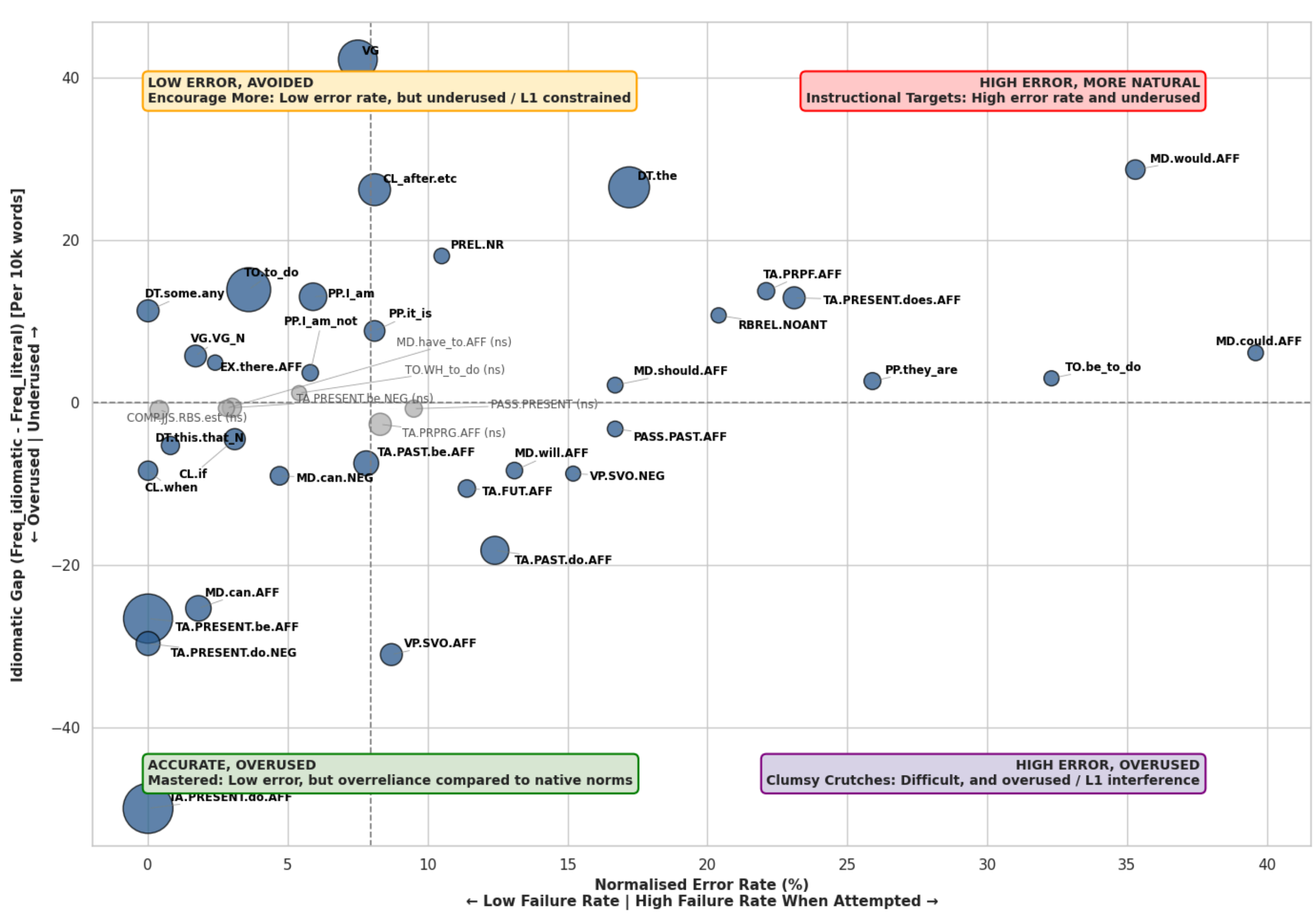

Figure 4: Grammatical target profile: Error rate (%) vs. idiomatic gap.

Grammar items that demonstrate a statistically significant shift ($p < 0.05$) in accuracy or idiomatic revision are rendered in color, whereas items that fail to reach significance on any metric are greyed out (ns). Dividing the feature space at the median error rate (8.0%) and a neutral idiomatic baseline (0.0) yields four distinct instructional quadrants (Figure 4), categorising targets based on whether learners struggle with structural execution, stylistic overuse, or avoidance. In the upper-right quadrant (*High Error, More Natural*), targets such as hypothetical modals (**MD.could.AFF**, **MD.would.AFF**), complex agreement and aspect (**TA.PRPF.AFF**, **TA.PRESENT.does.AFF**, **PP.they_are**), and definite articles (**DT.the**) exhibit elevated error rates alongside large positive idiomatic gaps. Because of their higher difficulty, subordinate clauses (**CL_after.etc**) and nonrestrictive relative pronouns (**PREL.NR**) also fall into this quadrant. Conversely, the upper-left quadrant (*Low Error, Avoided*) contains non-finite verb forms (**VG**, **TO.to_do**), first-person pronouns (**PP.I_am**), and determiners (**DT.some.any**), which exhibit low error rates when attempted but are substantially underrepresented relative to native revisions.

In contrast, features in the lower half of the feature space demonstrate negative idiomatic gaps, indicating reduction during native-like revision. The lower-left quadrant (*Accurate, Overused*) is dominated by basic simple present verbs (**TA.PRESENT.do.AFF**), present BE stems (**TA.PRESENT.be.AFF**), modal *can* (**MD.can.AFF**), and adverbial clauses (**CL.when**), which learners produce with high accuracy but rely on as repetitive default templates. Finally, the lower-right quadrant (*High Error, Overused*) contains narrative tenses such as simple past lexical verbs (**TA.PAST.do.AFF**), future forms (**TA.FUT.AFF**), past passives (**PASS.PAST.AFF**), and base Subject-Verb-Object patterns (**VP.SVO.AFF**), which present notable error rates while simultaneously being overproduced relative to native-like revisions.

# 5. Discussion

This study introduced and validated a framework for disentangling grammatical accuracy from native-like idiomaticity in Japanese EFL writing, revealing that these dimensions represent distinct developmental challenges with divergent pedagogical implications. Our findings demonstrate that treating all deviations from a native standard as homogeneous "errors" obscures crucial distinctions between structures that learners attempt but execute incorrectly (accuracy gap) and those they avoid altogether or overuse as default templates (idiomatic gap).

The accuracy gap findings corroborate prior work on L1 transfer in Japanese EFL writing. Given that these learners are just beginning their formal English language education, the prevalence of these gaps is unsurprising; the critical diagnostic contribution lies not in the gaps' existence, but in how their patterns differ across distinct grammatical features. The substantial increase in definite articles (+17.27 Δ, $p < .001$) following correction aligns with established patterns of article omission (Hinenoya & Lyster, 2015), but our data further reveal that articles also show a large idiomatic gap (+26.49 Δ), indicating that learners both err in article use and avoid them in natural contexts. This dual difficulty suggests that article instruction must address both grammatical form and pragmatic usage; a nuance lost in traditional error counts. Similarly, third-person singular *-s* (**TA.PRESENT.does.AFF**) exhibits significant gaps in both dimensions (+4.92 and +12.89 Δ respectively), consistent with Muroya's (2018) finding that Japanese learners find this morpheme disproportionately difficult due to its absence in the L1 morphosyntactic system. The modal system presents an instructive asymmetry: while *can* is heavily overused (-25.33 Δ, log ratio -1.04), reflecting L1 transfer from Japanese *dekiru*, *would* and *could* show both accuracy difficulties and avoidance patterns. This suggests that learners resort to a limited set of modal defaults while failing to develop the full modal system; a language developmental bottleneck that explicit instruction should address.

Perhaps more significant for pedagogy is the identification of structures that are grammatically accurate when attempted but substantially underused relative to native-like revisions. Nonrestrictive relative pronouns (**PREL.NR**) exhibit the largest proportional idiomatic gap (log ratio 3.86), suggesting that Japanese learners avoid post-nominal modification (e.g. "*the house, which I bought last year, is nearby*") and instead splits thoughts into separate sentences or put descriptors first, likely due to L1 pre-nominal relative clause structure (Strong et al., 2026). Present perfect (**TA.PRPF.AFF**) similarly shows significant underuse (+13.71 Δ, log ratio 1.51), reflecting the absence of a perfect aspect in Japanese and consequent reliance on simple past. These findings challenge the assumption that grammatical instruction alone suffices; these structures require extensive exposure and production practice to move from declarative knowledge to active usage.

The quadrant-based typology utilised for RQ3 provides a framework for instructional prioritisation. Structures in the upper-right quadrant (*High Error, More Natural*), including the definite article, third-person *-s*, *would*, present perfect, subordinate clauses, and nonrestrictive relatives, require both explicit grammar instruction and forced production practice to address this dual difficulty. Conversely, structures in the upper-left quadrant (*Low Error, Avoided*), such as non-finite verb forms, first-person pronouns, and determiners (*some/any*), demand awareness-raising and extensive exposure rather than further form-focused teaching (Nation, 1996). The lower-left quadrant (*Accurate, Overused*) contains default templates including simple present, present *BE*, *can*, and adverbial *when* clauses, where instruction should focus on providing lexical and syntactic alternatives. Finally, the lower-right quadrant (*High Error, Overused*), containing narrative tenses, future forms, past passives, and SVO patterns, requires attention to both grammatical execution and diversification away from these overused templates. This differentiation moves pedagogical intervention from intuition to evidence, enabling teachers to target specific deficits with appropriate methodologies.

## 5.1 Contributions

Theoretically, these results support the idea that first language influence operates at multiple levels: word structure, sentence structure, and meaning in context (Jarvis & Pavlenko, 2010). Accuracy gaps (e.g., with *-s* and articles) reflect the direct transfer of Japanese word-building rules into English. Idiomatic gaps (e.g., with relative clauses and present perfect), by contrast, reflect the transfer of larger sentence patterns and conceptual habits. Overuse patterns (simple present, SVO) further highlight a common developmental strategy: when unsure, learners fall back on simple Japanese-mapped structures as a safe default (Takahashi, 2016).

Methodologically, the use of LLM-generated layered correction and revisions proved effective in isolating these dimensions, though limitations warrant acknowledgment. The use of an open source LLM allows reproducibility, but manual validation was ad hoc and restricted to a subset of outputs. The rule-based CEFR-J grammar extractor, while avoiding LLM hallucinations, may fail to recognise certain error types or misinterpret non-standard forms; particularly when lexical errors obscure the underlying grammatical structure intended by the learner. Future work should incorporate native speaker correction and revisions, and multiple LLM comparisons to validate idiomatic judgments. Additionally, the diary-task context may have elicited specific patterns; examining argumentative or academic genres would reveal whether gaps persist across task types. Future studies should also extend this framework to other proficiency levels and L1 backgrounds to assess the generalisability of the gap profiles and distinguish developmental patterns from language-specific transfer effects.

For classroom practice, these findings offer actionable diagnostic profiles. Rather than providing undifferentiated grammar feedback, teachers can identify whether a student's difficulty stems from inaccurate execution, structural avoidance, or over-reliance on defaults. A student who produces accurate but overly simple prose requires different intervention than one who attempts complex structures with errors. The CEFR-J grammar extractor enables deterministic fine-grained profiling by proficiency level, allowing teachers to set developmentally appropriate targets.

## 6. Conclusion

This study operationalises a layered LLM-correction pipeline to quantify distinct accuracy and idiomatic gaps in EFL writing, moving beyond global error counts to diagnosis at the level of individual grammatical features. By mapping structures onto a two-dimensional typology—error rate versus idiomatic gap—the framework distinguishes overused default templates from avoided complex forms, enabling differentiated pedagogical responses. Anchoring this approach to the deterministic, CEFR-J-aligned grammar extractor ensures replicability and transparency. While further human validation of LLM outputs remains a necessary next step, the protocol offers a testable and scalable method for translating the accuracy vs. idiomaticity theoretical distinction into actionable, classroom-ready diagnostic profiles.

## 7. Acknowledgements

Many thanks to the English teacher and students who participated in this research, which was supported by Council for Science, 3rd SIP JPJ012347, and JSPS Grant-in-Aid for Scientific Research (B) JP23H01001, (A) JP23H00505, and KAKENHI Grant Number 26KJ1488.

## 8. Bibliographical References

## Appendix

Text correction and idiomatic revision prompts:

```
literal_correction_system_instruction =
    "You are an expert EFL literal
    proofreader. Your single task is to
    fix spelling, punctuation, and
    localised grammatical errors (such as
    missing subject-verb agreements,
    articles, or tense markers).
    CRITICAL: Do NOT alter sentence
    structure, change vocabulary choices
    to sound more elegant, combine
    sentences, or reorganise the syntax.
    Keep the student's phrasing and
    literal thought pattern exactly as it
    is, even if it sounds clumsy or
    unnatural. Output ONLY the corrected
    text without explanations or
    introductory text."

idiomatic_revision_system_instruction =
    f"You are an expert native English
    editor. Rewrite the student's text so
    it sounds completely natural, fluent,
    and idiomatic for a student answering
    this specific prompt:
    '{writing_prompt_text}'. Optimise
    vocabulary flow, use appropriate
    idioms, and smooth out awkward
    phrasing while keeping their core
    meaning WITHOUT adding or removing
    content. Output ONLY the revised
    text."
```